# Camera identification by grouping images from database, based on shared noise patterns

Teun Baar, Wiger van Houten, Zeno Geradts
Digital Technology and Biometrics department, Netherlands Forensic Institute, Laan van Ypenburg 6 2497 GB The Hague, The Netherlands

*Previous research showed that camera specific noise patterns, so-called PRNU-patterns, are extracted from images and related images could be found. In this particular research the focus is on grouping images from a database, based on a shared noise pattern as an identification method for cameras. Using the method as described in this article, groups of images, created using the same camera, could be linked from a large database of images. Using MATLAB programming, relevant image noise patterns are extracted from images much quicker than common methods by the use of faster noise extraction filters and improvements to reduce the calculation costs. Relating noise patterns, with a correlation above a certain threshold value, can quickly be matched. Hereby, from a database of images, groups of relating images could be linked and the method could be used to scan a large number of images for suspect noise patterns.*
PRNU link image noise pattern threshold value correlation database

## Introduction

To get more information about the origin of certain digital images, previous studies showed the ability to find the camera fingerprints on digital images. The fingerprint (so-called PRNU-patterns), are visible in the form of image noise. The noise is created because of little artifacts of the camera image sensor and could be extracted as shown by Lukas et al. [1].

Different studies [2, 3, 4] have shown that using this information, forensic scientist are able to match images and cameras based on fixed image noise caused by the sensor as an indicator for the camera used. The results may for example be of interest in cases of child abuse or pornography.

While previous work at the Netherlands Forensic Institute [5], focussed on low quality images/videos, showed the ability to combine images to get a clear noise pattern for investigation, the goal of the following research was to learn about the possibility to match single images to known noise patterns. Hereby one can think of scanning through a lot of images (online) to find correlating noise patterns (fingerprints). Therefore fast noise pattern extraction methods should be used and information about the reliability should be formed. The central part of the project is digital image noise patterns, which can be used as identification method for digital cameras. These so-called PRNU- patterns are the fingerprint of the intern camera chip on the digital image, caused by the non-uniformity of the chip.

## Using image noise patterns

When noise patterns are extracted from images, forensic scientists could use these patterns to compare to others and get information about the relation between those images. In the next situations noise pattern analyses could be used:
1. Determine if an image is made using the suspect camera
2. Determine if a group of pictures is made using the same camera
3. Determine groups of images created using the same camera, from a database of images

Currently, the first situation is one where already a lot is investigated. The noise pattern of single suspect images is compared to the noise pattern of images created using a suspect camera and other reference patterns from the same camera model. From these results can be concluded whether those suspect images are created using the suspect camera.

The image noise pattern (or PRNU), that is extracted from the images, is often the averaged result of multiple noise patterns (like shown by [5]), to form the camera specific noise. Like shown in Figure 1, when comparing PRNU patterns from cameras of the same type, matching camera noise patterns could be distinguished from others for most camera model as the correlation of the corresponding noise patterns is calculated.

Being able to suppress the consequences of periodic noise, for all cameras reliable correlation plots could be created. Where now images could be matched (eg. seen as created using the same camera) or mismatched, based on the correlation of the noise patterns.

## Extracting groups from databases

As described before, a possible usage of noise patterns could be finding images, which were created using the same camera from a huge database of images. The following 6 step process show how correlating noise patterns of images are put together to form groups as shown below:

i) The database is split in blocks of 50 images and for each of the blocks, the next steps ii) to v) will follow
ii) First a list is created for all images in the block and for each one the noise pattern is calculated.
iii) A random chosen image is defined as the first group with the PRNU- patterns defined as the noise pattern of this image.
iv) Each image noise pattern, correlating to the PRNU of this group above a chosen correlation threshold value, is added to the group and the PRNU- pattern is now defined as the average noise pattern of the containing images.
v) The previous step is repeated until no other images from the database correlate to the group. Hereafter from step iii) the process is repeated until all images of the database are put in groups.
vi) After groups are formed for each block, the correlation of the PRNU for groups from different blocks is calculated. Based on this value, groups are combined in case of a match.

In this final step vi), correlations based on 90 degrees of rotation between noise patterns are compared to a threshold value to determine possible matches. Also different tests on edited noise patterns could be covered in this step; like scaled noise patterns (for images on different resolution, but created using the same camera), what could be caused by digital zoom, cropped images or images with different resolutions are combined in the database.
For these steps the time to calculate each noise pattern from an image and the size of this noise pattern (for correlation calculation) determine the actual time the process takes. Therefore, methods and approximations to speed up the process are important.
Faster algorithms could also be used to screen a lot of images (on a website or stream of uploading images) on certain suspect noise patterns.

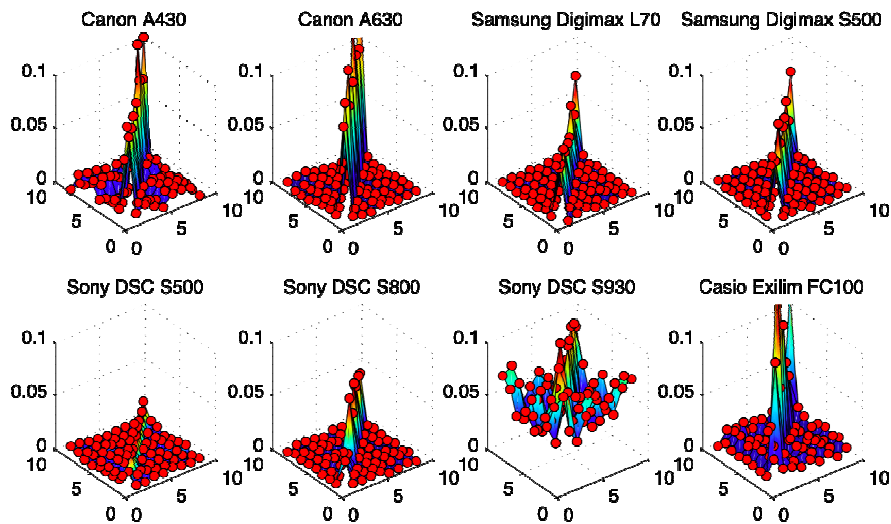

**Figure 1: Correlation plot for each tested camera model showing the correlation of each PRNU1 for camera 1-9 and PRNU2 for camera 1-9. PRNU1 and PRNU2 are calculated as the average noise pattern of 10 natural images from each camera. Thus correlations on the diagonal correspond to matching PRNUs (created using the same camera). The strange behavior of the Sony DSC S930 camera was caused by periodic noise in both the horizontal en vertical direction of the image at a frequency of 8 pixels, probably caused by the software to transform the sensor output to a JPG- compressed image.**

## Methods and approximations

Where different methods are available for extracting image noise patterns, they differ in accuracy and calculation speed. A reliable method like discussed by [1] (using a calculation in the 'Wavelet domain') could be used to calculate the noise pattern of an image. But the time to extract the noise pattern could be improved when using the Fourth Order Differential Filter proposed by [7], increasing the speed (by a factor of 8, as seen in the MATLAB programs) where this filter does not need the transformation to the Wavelet domain. When using a Second Order Differential Filter [6], the speed of the first mentioned method could even be increased (by a factor of 14), when an algorithm calculates the Second Order Differential of an image and subtract this from the original image to create a noise pattern.

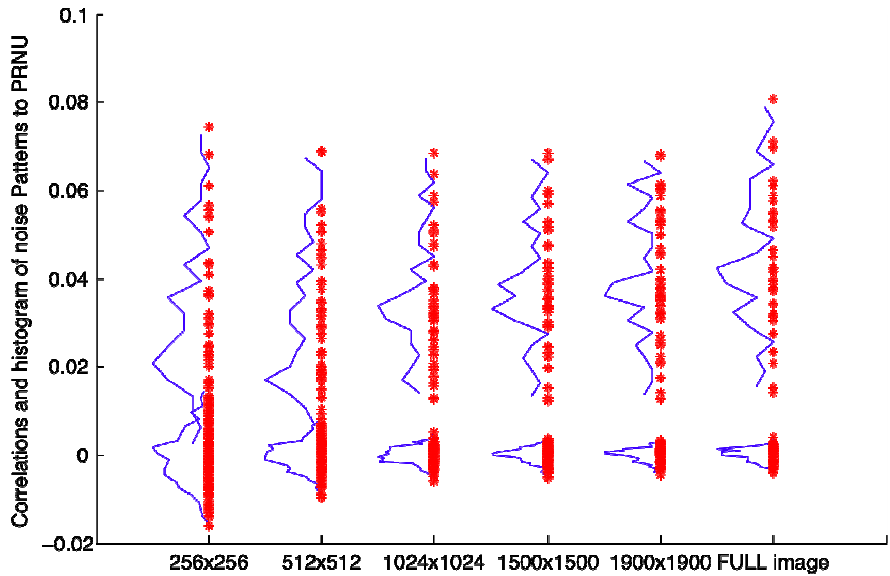

**Figure 2: Histogram (line) of correlation points (dots) for image noise patterns to PRNUs. Calculated for different image-regions as indicated on the x-axis. From the graph can be concluded that the performance of the camera identification raise with the used image size.**

Using over 3000 pictures created with 160 cameras of 8 different types, the performance of these methods are tested. The results show a small difference between the methods based on the used images. Noise patterns from flatfield images (often used as reference images) show the best results when the first method is used. Second and Fourth Order Differential filters showed best results in the case of natural images. The jpg-images that were used in the experiment where all natural images of original image size (order of 8 megapixel) for which the data was saved in the standard 3color RGB channels. Results for the natural images test can be seen in Figure i.

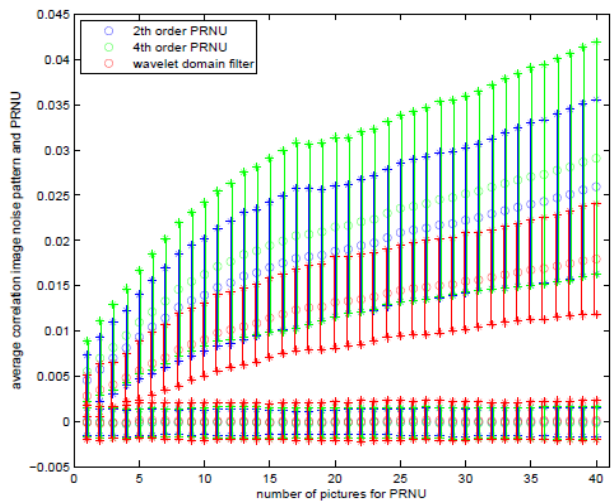

**Figure i. Average correlation of image noise patterns to PRNU patterns from the same camera (upper plots, matches) and different camera (lower plots, mismatches). The means and standard deviation are shown for the 3 methods.**

These results indicate that the differences between matches and mismatches are the largest for Fast Noise of the Fourth Order Differential filter.

Another test was done to see how well the different algorithms would perform when one natural image was compared to a set of 40 natural images. 6 cameras were used and per camera 8 natural photos were made.

Each photo was then compared to the set of 40 natural images. Results can be seen in Figures ii, iii and iv.

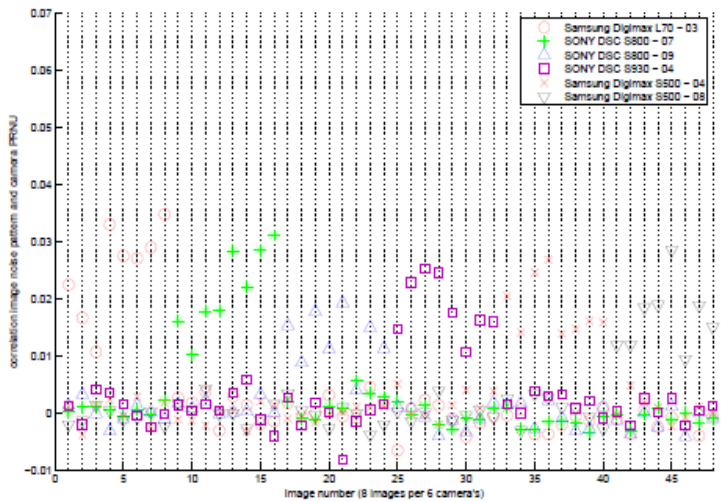

**Figure i: Plot of correlation between the noise pattern (of 8 images from 6 different cameras) and the PRNU (calculated as the avererage noise pattern of 40 images). The noise patterns are made using the wavelet domain filter.**

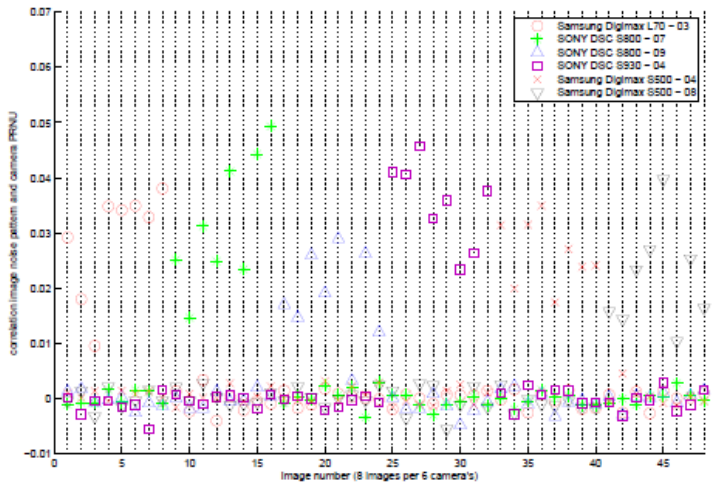

**Figure ii: Plot of correlation between the noise pattern (of 8 images from 6 different cameras) and the PRNU (calculated as the average noise pattern of 40 images). The noise patterns are made using the 2nd order method.**

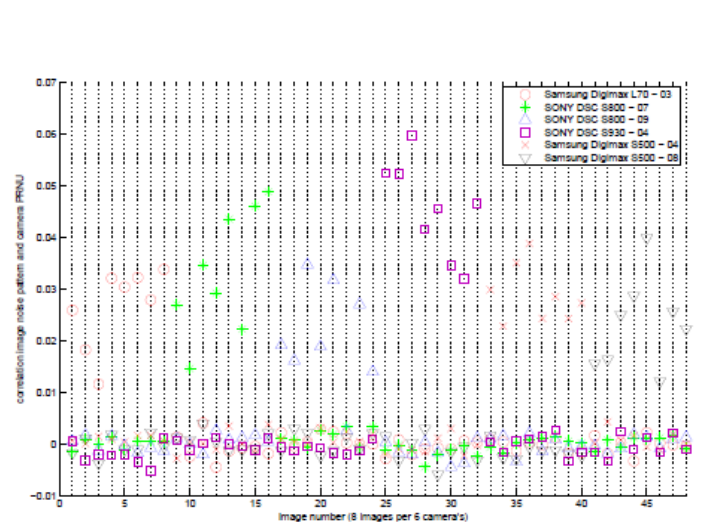

**Figure iii: Plot of correlation between the noise pattern (of 8 images from 6 different cameras) and the PRNU (calculated as the average noise pattern of 40 images). The noise patterns are made using the 2nd order method.**

The three algorithms seem to behave in pretty much the same way. For example, the correlation of the third image shows a dip compared to the first and second image for all the algorithms and image 22 resulted in a mismatch for each of the used methods. Again, the Fourth Order filter has the largest differences between matches and mismatches.

Because the calculation of the noise pattern is done per color channel, using only one instead of 3 could save extra time during the calculation. The influence of using the different color channels on the performance were tested and results show that using a combination of the 3 color channels shows the best results. Here all channels are added to a grey image (one channel) before the noise pattern is calculated. The size of the images influences the calculation time and reducing the image- size could therefore speed up the process. Using only a selection of the image results in a higher variance for matching and mismatching correlations and therefore a smaller distinction between the two. After some experiments, reducing the image size to 1024x1024 pixels still showed a clear distinction between matching and mismatching images, as shown in Figure 2. When the image size is reduced, only the pixels from the center of the image are used, because in this way rotations and scaled/shifted noise patterns can still be found. As shown in Figure 3, the suggested steps are to create a grey image of one color channel by adding the 3 color channels of the original image. Hereafter only the middle 1024x1024 pixels could be used for the calculation of the noise pattern to be compared to others (by calculation the correlation of possible matching noise patterns).

## Threshold value

To determine the distinction between matching and mismatching images, a threshold value is needed, for which the correlation of image noise patterns above this value is seen as matching images. This threshold value should be low enough for a high True Positive Rate (TPR), to find as most matching noise patterns. But the value should be high enough for a low False Positive Rate (FPR), not making too many errors (false combination). Therefore constant FPR rates (Equal Error Rates), is used to determine the threshold value, for example on a 1% error margin.

When a noise pattern was compared to a growing noise pattern (as the average of a growing number of image noise patterns), the threshold value should increase too, for a constant FPR (error). This is caused by the fact that for growing PRNU-patterns other non-camera specific noise filters out and correlation values increase.

Threshold values are determined for different combinations of PRNU-patterns and the results shown in Figure 4. The results of this varying threshold value (that was also determined for the combination noise patterns, both created using multiple images) were used in the previously discussed method to link groups in a large database.

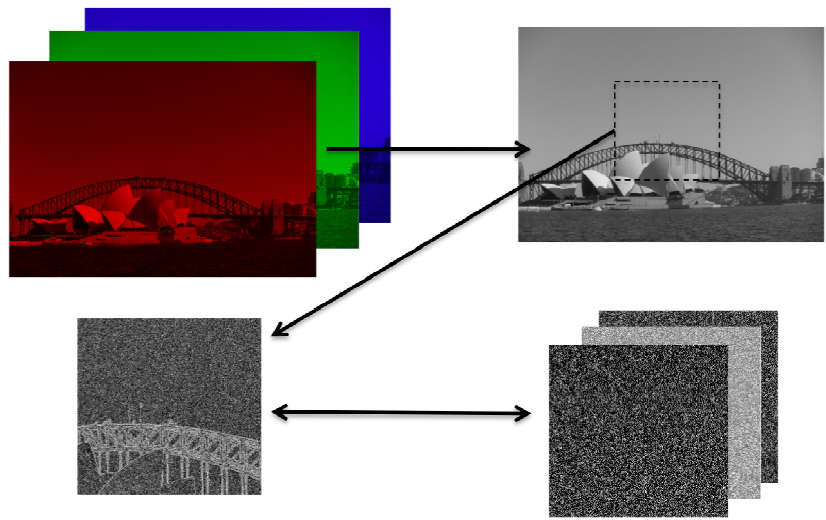

**Figure 3: Steps in the process of extracting noise patterns, where first the image is loaded and 3 color channels are added to a grey image. Hereafter, an imagepart of 1024x1024 pixels from the center is used to calculate the noise pattern (using a 2nd order differential filter). The correlation of the obtained noise pattern and reference patterns is calculated, to conclude if original images correlate.**

To find matching images based on correlating noise patterns, threshold values are created using a sample database. For a noise pattern A based on a-images and noise pattern B based on b-images; images A and B are concluded to match (created using the same camera) if corr2(A(a),B(b)) > threshold(a,b). Here the threshold values threshold(a,b) are created using a sample database of image noise patterns of the same size and using the same noise extraction method. To create the threshold(a,b) value an error margin r (like for example r = 1%) is determined as the value for which r% of the correlation between non-matching noise patterns from the sample database exceeds this value.

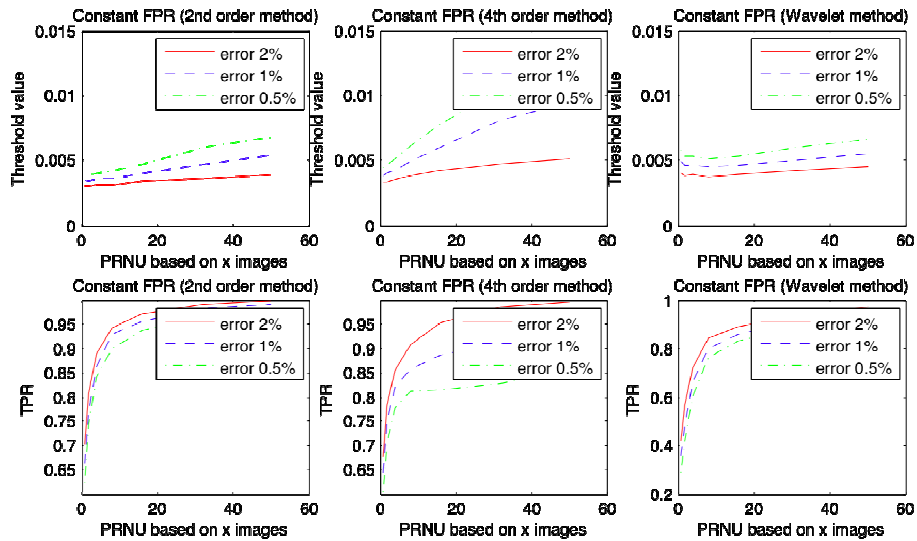

**Figure 4: Plots of threshold value at constant FPR (error) for increasing number of images to create PRNU. The correlations are calculated between the PRNU patterns and the noise patterns of single images of summation of the color channels and a picture size of 1024x1024 pixels.**

## Conclusions

The Fast Noise of the Fourth Order and the Fast Noise of the Second order Differential filters perform just as well or maybe even better than the Wavelet based filter. The advantage is that the Fast Noise filters are much faster than the Wavelet filter.

Using the methods described in Figure 3, groups were extracted from different databases, grouping images based on the camera used to create the original images. The calculation (for databases up to 500 images) was done in about 30 minutes, using above described methods and approximations in a MATLAB environment. The output showed groups that were indeed created using the same camera, plus some single images that could not be matched. The number of groups that were found was in the order of (often 1.5 times) the number of cameras used to create the database. Those experiments were done using the available pictures from the previous tests. Thereby images from other locations like pictures from the image website Flickr.com, were combined in a large folder and could easily be linked by this method.

Although, because of the small error margin, possible mismatches could occur while extracting groups from an image database, the results show this method could certainly be used. The results could be used in forensic science to find the source of images from a large folder. If necessary, selected groups could further be analyzed by using other methods or without approximations. Like the full size of the image could be used for the pattern noise calculation and calculate the correlation for certain groups.

Overall these findings show the possibility to investigate single images on corresponding noise pattern to quickly find relating images from large databases.

## References

[1] Lukas J.,Fridrich J., Goljan M.: Digital camera identification from sensor pattern noise Proc. SPIE Electronic Imaging, Image and Video Communication and Processing, San Jose, California, pp. 249-260, (2005)
[2] Chen M., Fridrich J.,Goljan M.,Lukas J.: Source Digital Camcorder Identification Using Sensor Photo Response Non-Uniformity Proc. SPIE, vol. 6505, 65051 G (2008)
[3] Khanna N., Mikkilineni A.K., Chiu G.T.C., Allebach J.P., Delp E.J. Forensic Classification of Imaging Sensor Types presented at the SPIE Int. Conf. Security, (2007)
[4] Celiktutan O., Sankur B., Avcibas, Y. Blind Identification of Source Cell- Phone Model IEEE transactions on information forensics and security, vol. 3, no. 3, (2008)
[5] Houten W. van, Geradts Z.: Source video camera identification for multiply compressed videos originating from youtube Digital Investigation, Issues 1- 2, pages 48-60, September (2008)
[6] Rudin L.I., Osher S., Fatemi E.: Nonlinear total variation based noise removal algorithms Physica D, vol 60, pp 259-268, (1992)
[7] Lysaker M., Lundervold A., Tai Xue-Cheng: Noise removal using fourth- order partial differential equation with applications to medical magnetic resonance images in space and time IEEE transactions on image processing, vol 12, no 12, (2008)